\documentclass{article}
 \PassOptionsToPackage{numbers, compress,sort}{natbib}
 \usepackage[final]{neurips_2021}
\usepackage[ruled,vlined]{algorithm2e}
\usepackage{algorithmic}
\usepackage[utf8]{inputenc} 
\usepackage[T1]{fontenc} 
\usepackage[hyperfootnotes=false]{hyperref} 
\usepackage{url} 
\usepackage{booktabs} 
\usepackage{amsfonts} 
\usepackage{nicefrac} 
\usepackage{microtype} 
\usepackage[utf8]{inputenc} 
\usepackage[T1]{fontenc} 
\usepackage{hyperref} 
\usepackage{url} 
\usepackage{booktabs} 
\usepackage{amsfonts} 
\usepackage{nicefrac} 
\usepackage{microtype} 
\usepackage{graphicx}
\usepackage[table]{xcolor}
\usepackage{makecell}
\usepackage{gensymb}
\usepackage{subcaption}
\usepackage{amsthm}
\usepackage{mathrsfs}
\usepackage{amsmath}
\usepackage{amssymb}
\usepackage{multirow}
\usepackage{mathtools}
\usepackage{wrapfig}
\usepackage{soul}
\makeatletter
\def\BState{\State\hskip-\ALG@thistlm}
\makeatother
\newcommand{\cdotv}{\boldsymbol{\cdot}}

\usepackage[textwidth=1.2in,textsize=tiny]{todonotes}

\newcommand{\ba}[1]{\begin{align}#1\end{align}}

\newcommand{\minus}{\scalebox{0.5}[1.0]{$-$}}

\DeclareMathOperator*{\argmin}{argmin}

\usepackage{stackengine}

\makeatletter
\newcommand{\distas}[1]{\mathbin{\overset{#1}{\kern\z@\sim}}}%

\newcommand{\beqs}{\vspace{0mm}\begin{eqnarray}}
\newcommand{\eeqs}{\vspace{0mm}\end{eqnarray}}
\newcommand{\barr}{\begin{array}}
\newcommand{\earr}{\end{array}}

\newcommand{\xv}{\boldsymbol{x}}
\newcommand{\yv}{\boldsymbol{y}}

\newcommand{\thetav}{\boldsymbol{\theta}}

\newcommand{\given}{\,|\,}

\usepackage[symbol]{footmisc}

\renewcommand{\thefootnote}{\fnsymbol{footnote}}
\newcommand{\ours}{{AA}}
\title{
Alignment Attention by Matching\\ Key and Query Distributions} 
\author{Shujian Zhang \quad Xinjie Fan \quad Huangjie Zheng \quad Korawat Tanwisuth \quad Mingyuan Zhou\\
The University of Texas at Austin
\\

\texttt{\{szhang19, xfan, huangjie.zheng, korawat.tanwisuth\}@utexas.edu}
\\\texttt{mingyuan.zhou@mccombs.utexas.edu}
}

\begin{document}

\maketitle

\begin{abstract}

The neural attention mechanism has been incorporated into deep neural networks to achieve state-of-the-art performance in various domains. Most such models use multi-head self-attention which is appealing for the ability to attend to information from different perspectives. This paper introduces alignment attention that explicitly encourages self-attention to match the distributions of the key and query within each head. The resulting alignment attention networks can be optimized as an unsupervised regularization in the existing attention framework. It is simple to convert any models with self-attention, including pre-trained ones, to the proposed alignment attention. On a variety of language understanding tasks, we show the effectiveness of our method in accuracy, uncertainty estimation, generalization across domains, and robustness to adversarial attacks. We further demonstrate the general applicability of our approach on graph attention and visual question answering, showing the great potential of incorporating our alignment method into various attention-related tasks.{\let\thefootnote\relax\footnote{The code is available at \url{https://github.com/szhang42/alignment_attention}}}


\end{abstract}

\section{Introduction}
Attention-based mechanisms aggregate features with learnable weights to introduce useful inductive biases for sequence models  \cite{sutskever2014sequence,bahdanau2015neural}. Since the introduction of the self-attention based Transformer \cite{vaswani2017attention}, attention has become the foundation for many state-of-the-art models.
Exploiting its computational efficiency and scalability,
it has been used to 
train unprecedented large models on big datasets \citep{devlin2018bert}.
Large attention-based 
models have been demonstrating their ability to learn good representations in an unsupervised manner and benefit downstream analysis, 
with tremendous success in various natural language processing (NLP) \citep{devlin2018bert,lan2019albert, liu2019roberta,joshi2020spanbert,radford2018improving,yang2019xlnet}, compute vision \citep{dosovitskiy2020image,chen2020generative}, and multi-modal learning tasks \citep{chen2019uniter,lu2019vilbert}. 

Attention networks, including multi-head attention, are being effectively utilized to capture the correlations between each pair of input tokens through individual or multiple attention functions.  More specifically, in a self-attention layer with $H$ heads, assuming that the output of the previous layer consists of $n$ tokens, each of which is represented as a  feature vector of dimension $d_{model}=d\cdotv H$, then each token feature vector will be transformed by a $d_{model}\times d$ query projection matrix into a query feature vector, by a $d_{model}\times d$ key projection matrix into a key feature vector, and by a $d_{model}\times d$ value projection matrix into a value feature vector. The inner products of the $i$th query feature vector with all $n$ key feature vectors are then fed through a softmax function to define the relative weights of the $n$ keys to that query, which are used to aggregate the $n$ value vectors into the vector representation of the $i$th word token in a head. 

Although such networks are simple to optimize and intuitive to understand, how the key and query projection matrices should differ from each other has not been well-studied and understood. It is thus unclear whether they would result in well-controlled 
interactions between the keys and queries.
In particular, ignoring the dependence between the $n$ token feature vectors, we can view the output of the previous layer as an empirical distribution supported on $n$ points in $\mathbb{R}^{d_{model}}$. In each head, this empirical distribution is transformed by the query and key projection matrices into a query empirical distribution and a key empirical distribution, respectively, each supported on $n$ points at the same feature space in $\mathbb{R}^{d}$. 
Since at each head two different projections matrices  are used to project the input,
the distributions of key and query will be different.
Intuitively, 
if these two distributions are clearly misaligned
with each other, then the query and key of a token, whose input feature resides in a region with lower probabilities, may have increased risk of being pushed further away from each other in the shared projection space. 






This paper proposes alignment attention, which regularizes the query and key projection matrices  at each self-attention layer, by matching the empirical distributions of the query and key feature vectors. 
We focus on within-head alignment between empirical distributions and present three different 
 options for distribution matching. In our framework, alignment attention, built as an unsupervised approach to match the query and key distributions, is trained jointly to maximize a combination of data likelihood and distribution agreement.
This efficient architecture design enables us to easily add alignment loss to
convert existing self-attention networks, including pre-trained ones, into alignment attention.
Meanwhile, it  naturally shares parameters and computation 
with the self-attention networks, allowing an end-to-end training. 

With a generic architecture, alignment attention can convert any existing soft attention models, including pre-trained ones, 
while maintaining the inherent advantages of conventional attention, such as  efficiency and being simple to optimize. 
The proposed method boosts the performance while 
remaining efficient in memory and computation cost.
Our experiments show that the proposed alignment attention method outperforms state-of-the-art self-attentions in a wide variety of settings, including natural language understanding tasks, graph attention network, and visual question answering, in terms of accuracy and uncertainty estimation.
We further demonstrate that alignment attention achieves strong performance in domain generalization and adversarial robustness.


\section{Alignment attention}

We introduce a general recipe for alignment attention: (a) build the alignment to match the key and query distributions within each head, (b) develop efficient distribution matching methods, and (c) leverage existing attention structure and optimize the model in an end-to-end manner.
The resulting architecture can be efficiently learned with existing self-attention networks.

\subsection{Attention modules}
Attention uses keys and queries to obtain soft attention weights $W$, which are then used to 
aggregate the values to obtain the output features. Consider $n$ key-value pairs with a key matrix $K\in \mathbb{R}^{n\times d_k}$, a value matrix $V\in \mathbb{R}^{n\times d_v}$, and $m$ queries $Q \in \mathbb{R}^{m\times d_k}$, where in general the dimensions of queries and keys are equal. The scaled product between key and query \citep{vaswani2017attention} is: $\Phi = f_\text{dot}\text(Q,K) = QK^T/\sqrt{d_k}\in \mathbb{R}^{m\times n}$.
Alternative 
choices include dot-product \cite{vaswani2017attention,devlin2018bert} and
additive attention \cite{xu2015show,rennie2017self,bahdanau2014neural}.
Attention weights $W$ is defined as the softmax output of $\Phi$: $W=\text{softmax}(\Phi)$, where 
$W_{i,j} = \frac{\exp (\Phi_{i,j}) }{ \sum_{j'=1}^n \exp (\Phi_{i,j'})}$ 
represents the importance of the $j$th key to the $i$th query learned by the neural networks.

The multi-head attention, first proposed in Transformer \cite{vaswani2017attention}, 
projects the queries, keys, and values into $H$ subspaces with $H$ different learnable linear projections. These projections are performed in parallel and then concatenated into a single latent representation. 
At the $l$th self-attention layer,
we can obtain attention weight $W^{l,h} = \text{softmax}(f(Q^{l,h},K^{l,h}))$, where $Q^{l,h} = Q^{l}M^{l,h}_Q$, $K^{l,h}=K^{l}M^{l,h}_K$, and $V^{l,h} = V^{l}M^{l,h}_V$ for $h=1,...,H$, with $M$ denoting the parametric matrices to learn. The attention results from all heads are then
concatenated into the layer output as $O^{l} = [W^{l,1}V^{l,1}, ..., W^{l,H}V^{l,H}]$.  


\subsection{Alignment attention}
Self-attention allows the model to attend to the information from each representation
subspace at each position \cite{li2018multi}. To encourage different attention heads to indeed capture
distinct features, most previous studies focus on the disagreement regularization to
explicitly encourage the diversity among
attention heads \cite{li2018multi}.
By contrast, 
our focus is on exploiting how to make the key and query distributions better interact with each other in the latent space. 
We propose agreement matching to encourage the distributions of key and query over different tokens to be consistent 
within each head.
Given a source input $\xv$ and its output $\yv$, a neural attention model is trained to maximize the conditional probability of $\yv$ given $\xv$ over a training corpus. We introduce  distribution alignment, an auxiliary regularization term in order to encourage the alignment between the learned key and query distributions. Considering a supervised learning problem with training data $\mathcal{D}:=\{\xv_i, \yv_i\}_{i=1}^N$, the likelihood parameterized by~$\thetav$ is denoted by 
$p_{\thetav}(\yv_i\given\xv_i)$. 
For notational convenience, below we drop the data index $i$. The whole model is differentiable to directly maximize the likelihood. Formally, the training objective with alignment attention is
expressed as:
%
%
\ba{\mathcal{L}(\xv, \yv)=\underbrace{\log {p_\theta}(\yv \mid \xv)}_{\text {likelihood }}+\lambda * \underbrace{ \mathcal{L_{\text{Align}}}(Q, K)}_{ \text {alignment }},
\label{eq:overall_loss_fuction}}
where 
$\lambda$ is the alignment weight 
\cite{berthelot2019mixmatch, zhang2021knowing, duan2021topicnet, zhang2021learning}.
The auxiliary regularization term $\mathcal{L_{\text{Align}}}$ guides the distribution matching between key ($K$) and query ($Q$).

The proposed alignment provides a sample-and-head-dependent matching between the key and query. Assuming $Q$ and $K$ have the same batch size and number of heads; $Q$ is of dimension 
$[B, H, n, d_q]$, where $B$ represents the batch size, $H$ the number of heads, $n$ the number of queries,
 and $d_q$ the hidden dimension within a head; and $K$ is of dimension $[B, H, m, d_k]$. 
Focused on the self-attention networks, we assume $n = m =w$ and $d_q = d_k = d $ in our alignment attention. 
We use
$Q, K$ 
to calculate the point-to-point difference from the query to key for each head at each sample, resulting in a tensor with dimension $[B, H, w, w]$.
At each of the $B$ samples of the minibatch and each of the $H$ heads, the training objective is to minimize the expected difference between the empirical  distributions of query and key, both of which are supported on a set of $w$ query/key features in $\mathbb{R}^d$ (see Figure \ref{fig:trans}).
This flexible alignment method 
could be conveniently deployed into a single-head or multi-head attention mechanism.

\begin{figure*}[t] 
 \centering
 \includegraphics[width=12.0cm]{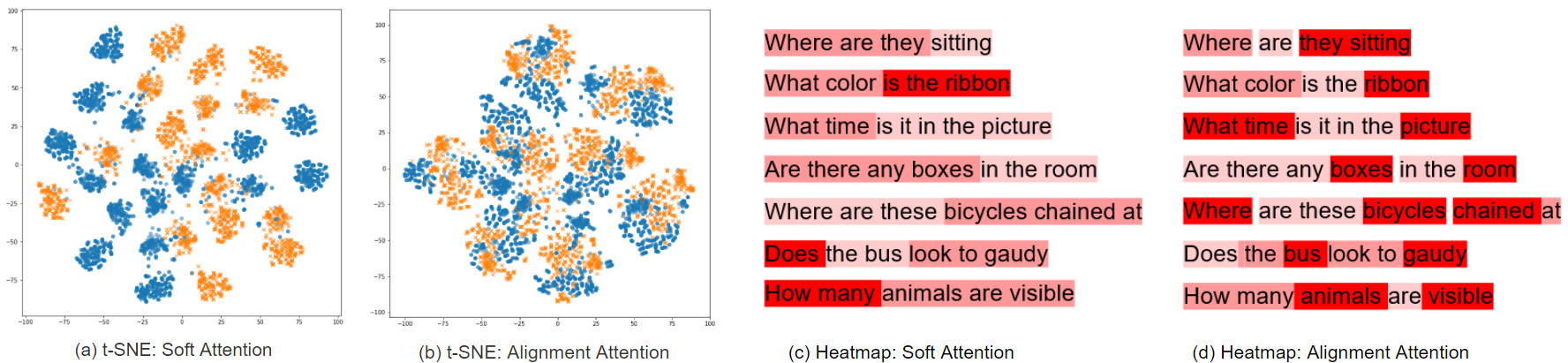}
 \caption{\small  T-SNE visualizations of (a) soft attention and (b) alignment attention with bidirectional conditional transport. In (a) and (b), we visualize the both key distribution and the query distribution. The blue dots and orange diamonds represent the features from the key distribution and these from the query distribution, respectively. We show attention heatmaps of a VQA data in (c) and (d). It includes overall attention of seven examples' matrix embedding by summing up the attention weight vectors, with darker color denoting higher attention probability.
 } 
 \label{fig:att_vis_explanation} \vspace{-4mm}
\end{figure*}

With the alignment attention, the resulting key and query distributions should be close to each other
 (see t-SNE plots \citep{van2008visualizing, duan2021sawtooth}in Figure \ref{fig:att_vis_explanation}).
We also visualize the attention weight from both soft and alignment attentions in Figure \ref{fig:att_vis_explanation}. It is clear that while many semantically and sentimentally  important words and their combinations, such as ``where, they, sitting,'' ``color, ribbon,'' ``what, time, picture,'' ``boxes, room,'' ``where, bicycles, chained,'' ``bus, gaudy,'' and ``animals, visible,'' are overlooked by vanilla soft attention, they are appropriately highlighted by the proposed alignment attention.  
\subsection{Alignment methods} \label{sec:options_for_align}




To align the key and query distributions, we need a method that can quantify the difference between two distributions given their empirical samples. 
Under this requirement, we consider three different distribution matching methods, including the discriminator-based adversarial training \cite{goodfellow2014generative}, which is 
directly related to the Jensen--Shannon (JS)
divergence \cite{lin1991divergence}, the Wasserstein distance  
in its primal form, which can be defined by solving an optimal
transport problem 
\cite{villani2008optimal,peyre2019computational},  and bi-directional conditional transport \cite{zheng2021act}, 
which is developed by exploiting both the chain rule and Bayes' rule to quantify the difference between two probability distributions.

\subsubsection{Adversarial training-based alignment}\label{sec:GAN} Adversarial training, the key part to enable GANs \citep{goodfellow2014generative}, has been successfully exploited to minimize the distributional discrepancy \cite{ganin2016domain, tzeng2017adversarial}.
Under a minimax two-player framework, the  discriminator $D$ is trained to distinguish the distributions of key and query, while both the query and key projection matrices, $M_Q$ and $M_K$, 
are treated as the generator $G$ and trained to confuse $D$. Here we consider a discriminator $D$ that is shared across the $H$ heads for alignment.
Denote the empirical distributions of query and key as $p_{Q}$ and $p_{K}$, respectively.
For a sample of $w$ tokens and at one of the $H$ heads, 
we have $p_{Q}=\sum_{i=1}^w \frac{1}{w}\delta_{q_i}$ and $p_{K}=\sum_{j=1}^w \frac{1}{w}\delta_{k_j}$, where $q_i$ and $k_i$ are projected from the $i$th token's input feature via the query and key projection matrices at that head, respectively. Thus the alignment loss 
is the sum over $H$ different head-dependent loss, each of which 
on a sample can be expressed as (we drop the head index for notational simplicity)
\ba{
\min _{G} \max _{D} \mathcal{L}_\mathrm{Align-GAN}(Q, K):=\mathbb{E}_{{q} \sim p_{Q}}[\log D({q})]+\mathbb{E}_{{k} \sim p_{K}}[\log (1-D({k}))].
}
%
%
In summary, the discriminator is trained to maximize the alignment loss, and the generator, $i.e.$, the query and key projection matrices, is trained to minimize the alignment loss, so that the key and query distributions are learned adversarially to align with each other.


\paragraph{Discriminator-based modules.} 

We leverage the highway architecture \cite{srivastava2015training} to construct the discriminator. 
With a feature map $X\in \{Q, K\}$,
the highway network first computes a transform gate $\tau$, as $
\tau=\sigma\left(\Phi_\tau(X)\right),$ where $\Phi_\tau$ is a fully-connected layer and $\sigma$ is the sigmoid activation function.
Then, the output of the highway network is, ${I_X}=\tau \odot \text{ReLU}(\Phi_h(X))+({1}-{\tau}) \odot X
,$
where
$\Phi_h$ is another linear layer followed by ReLU activation and $\odot$ denotes the element-wise multiplication. 
We further apply a two-layer MLP to obtain the classifier probability, $
D(X)=\sigma(\Phi_{2}\left(F_{\mathrm{NL}}\left(\Phi_{1}\left({I_X})\right)\right)\right))$, where $\Phi_{1}$ and $\Phi_{2}$ are fully-connected layers connected by $F_{\mathrm{NL}}$, a leaky ReLU activation function.
To optimize the discriminator-based modules, instead of alternately
updating the adversaries, like in GAN \citep{goodfellow2014generative}, we use the gradient-reversal
layer~\citep{ganin2015unsupervised} to jointly optimize all the components.

\begin{figure}[tp] \vspace{-2mm}
\centering
\includegraphics[width=14.0cm]{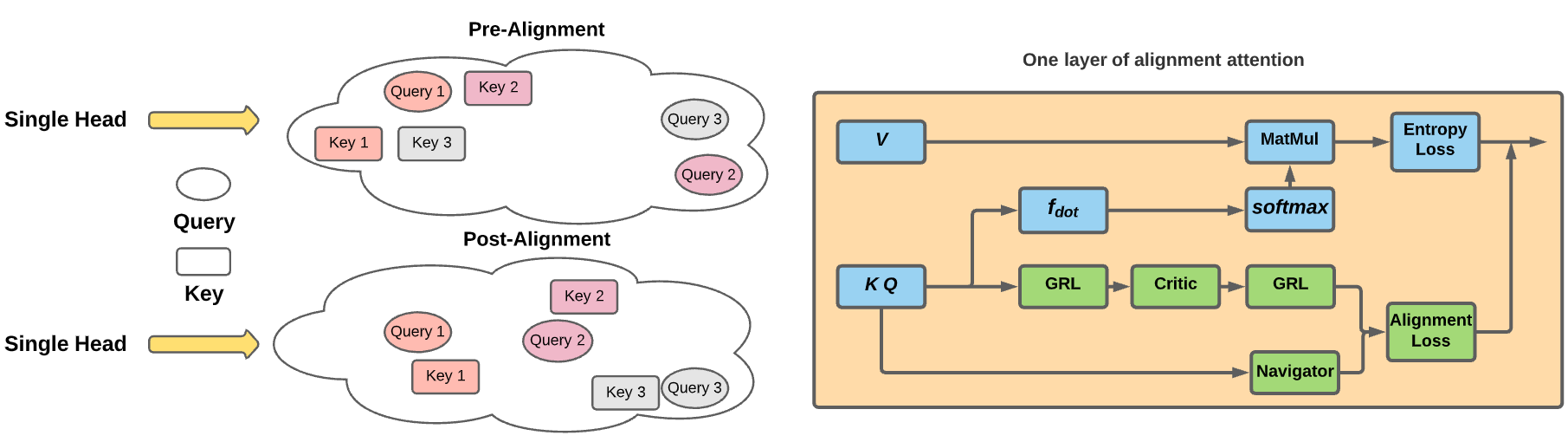}  \vspace{-2mm}
\caption{\small On the left, we visualize the alignment attention.
The ellipse represents query and the rectangle represents the key. On the right, a demonstration of the difference and similarity between vanilla soft 
attention and our alignment attention. Alignment attention (in green) shares the same architecture as soft attention before obtaining key, query, and value. Then alignment attention adds the alignment structure to perform distribution matching, where GRL represents the gradient reversal layer. 
} 
\label{fig:trans} \vspace{-5mm}
\end{figure}

\subsubsection{Optimal transport-based alignment}  An alternative to the adversarial training-based alignment is to 
consider optimal transport (OT) \cite{villani2008optimal}, which has been widely used 
for distribution matching.  
In OT, the transport plan $\Pi(p_Q,p_K)$ is the set of all possible joint distributions of query and key, whose marginals over the query and key are $p_K$ and $p_Q$, respectively. We define the OT-based alignment cost as
\ba{\mathcal{L}_\mathrm{Align-OT}(Q, K) = \min_{\pi\in\Pi(p_Q,p_K)}\mathbb{E}_{q,k \sim \pi}\left[ c(q,k) \right],}
where  $c(\cdot, \cdot)$ is the transport cost between two points.
For discrete $p_Q$ and $p_K$, fining the optimal transport plan often involves solving a computationally expensive linear programming problem.
It is also known in the literature that OT is sensitive to the outliers in $p_Q$ and $p_K$ due to the two marginal constraints \cite{balaji2020robust}. To reduce the computation burden, we consider an entropy-regularized OT, which allows the OT problem to be 
solved iteratively using the Sinkhorn iterations \cite{cuturi2013sinkhorn}. 

\paragraph{Optimal transport-based modules.} 
We use the Sinkhorn algorithm \cite{cuturi2013sinkhorn,feydy2019interpolating} to estimate the transport plan, defined as $\pi_{\epsilon} = \underset{\pi}\argmin \sum_{i=1}^w \sum_{j=1}^w (c(q_i, k_j) \cdot \pi(q_i, k_j) - \epsilon \pi(q_i, k_j) \log \pi(q_i, k_j))$, where $\epsilon = 0.01$. The OT-alignment cost now becomes $\mathcal{L}_\mathrm{Align-OT} = \sum_{i=1}^w \sum_{j=1}^w (c(q_i, k_j)  \cdot \pi_{\epsilon}(q_i, k_j))$.

\subsubsection{Bidirectional conditional transport-based alignment}
Instead of requiring $\pi\in\Pi(p_Q,p_K)$ as in OT, we follow \citet{zheng2021act} to consider probabilistic bidirectional conditional transport (CT), which exploits both the chain rule and Bayes' theorem to constrain the joint $\pi$ with both  $p_Q=\sum_{i=1}^w \frac{1}{w} \delta_{q_i}$ and $p_K=\sum_{j=1}^w \frac{1}{w} \delta_{k_j}$ in two different ways. First, we consider a query-to-key CT that define $\pi$ as $\pi(q_i,k_j)=p_Q(q_i)\pi_{K}(k_j\given q_i)$, where $\pi_{K}(k_j\given q_i)$ is a conditional distribution defined as $\pi_{K}(k_j\given q_i)=\frac{p_K(k_j)\exp (\tau_\phi(k_j)^T \tau_\phi(q_i)) }{ \sum_{j'=1}^w p_K(k_{j'})\exp (\tau_\phi(k_{j'})^T \tau_\phi(q_i))}$ and $\tau_\phi(\cdotv)$ is a neural network based transformation parameterized by $\phi$. Second, we consider a key-to-query CT that defines $\pi(q_i,k_j)=p_K(k_j)\pi_{Q}(q_i\given k_j)$, where $\pi_Q(q_i\given k_j) =\frac{p_Q(q_i)\exp (\tau_\phi(k_j)^T \tau_\phi(q_i)) }{ \sum_{i'=1}^w p_Q(q_{i'})\exp (\tau_\phi(k_{j})^T \tau_\phi(q_{i'}))}$. Third, we define a point-to-point cost as $c_{\eta}(q,k)=1 -\frac{\tau_\eta (k)^T\tau_\eta(q)}{||\tau_\eta(k)||_2||\tau_\eta(q)||_2},$
where $\tau_\eta(\cdotv)$ is a neural network based ``critic'' function whose parameter $\eta$ will be adversarially learned. Combing them together leads to the bidirectional CT-based alignment loss as

\ba{{\mathcal{L}_\mathrm{Align-CT}(Q, K) =  \textstyle\frac{1}{2} \mathbb{E}_{{q} \sim p_{Q},~k \sim \pi_{K}(\cdotv\given q)}[ c_\eta(q,k) ]+\textstyle\frac{1}{2} \mathbb{E}_{{k} \sim p_{K},~q \sim \pi_{Q}(\cdotv\given k)}[ c_\eta(q,k)].} \label{eq:act_loss_fuction}}

Compared to OT, bidirectional CT is able to efficiently model the alignment with less computation. 
The structure of alignment attention with transport-based methods is presented in Fig.~\ref{fig:trans}.

\paragraph{CT-based modules.} The critic $\tau_{\eta}(\cdotv)$ is structured similarly as the discriminator in Section~\ref{sec:GAN}. It projects the input data onto a vector space instead of outputting logits for binary classification. 
For $\tau_{\phi}(\cdotv)$, we use a two-layer MLP network. We optimize the query and key projection matrices and $\phi$ to minimize $\mathcal{L}_\mathrm{Align-CT}(Q, K)$ in \eqref{eq:act_loss_fuction}, and optimize $\eta$ to maximize it. The gradient-reversal
layer~\citep{ganin2015unsupervised} is also used to optimize the critic adversarially.

\section{Related work} 

\textbf{Alignment Learning.}
\citet{liang2006alignment} first assign agreement terms for jointly
training word alignment in phrase-based statistic
machine translation. 
The general bidirectional sequence alignment models with model
inevitability regularization are then proposed by \citet{levinboim2015model}. Recently, 
the alignment is studied in the multi-head attention model based on the Transformer architecture, where  alignment extraction is improved by augmenting an additional alignment head to the multi-head source-to-target attention component \cite{alkhouli2018alignment}. Our proposed alignment attention adopts the alignment idea and matches the distributions of query and key. This general and efficient framework gives us the flexibility to better model attention weights. The domain generalization ability and adversarial robustness of alignment attention are also studied. 

\textbf{Distribution Matching.} Distribution matching is a fundamental problem in statistics and machine learning \cite{murphy2012machine}. The widely used distances include  the Kullback–Leibler (KL) divergence \cite{kullback1951information} and Jensen–Shannon (JS)
divergence \cite{lin1991divergence}. GANs \cite{goodfellow2014generative} are proposed with the adversarial-based objectives. Due to the inherent advantages of allowing the two distributions to have non-overlapping supports \cite{arjovsky2017wasserstein, bellemare2017cramer}, the Wasserstein distance 
from the optimal transport problem is also used for defining the transport cost \cite{villani2008optimal, peyre2019computational}. In our alignment attention, we consider discriminator-based and transport-based methods as alignment methods. Based on the alignment methods, we apply alignment attention to the widely used attention models and leverage the existing efficient attention architecture to build entire alignment attention network.

\section{Experiments} \label{sec:experiemental_section}

Our method can be incorporated into any self-attention based models. To exam its effectiveness and general applicability, we apply alignment attention to a diverse set of tasks, including language understanding, graph attention, and visual question answering. Furthermore,  the model's generalization across domains and robustness towards adversarial attacks are studied on language tasks. 
We study a variety of state-of-the-art models for these tasks including ALBERT \cite{lan2019albert}, BERT \cite{devlin2018bert}, and RoBERTa \cite{liu2019roberta}. Below we present the main experimental settings and results, with more training and hyperparameter details provided in Appendix \ref{sec:app_exp}.

\subsection{Alignment attention in natural language understanding}\label{sec:bert}

Since the self-attention based Transformer was proposed, it has been widely used in pretrained models for NLP and related areas, achieving 
state-of-the-art results on various downstream tasks. However, training a state-of-the-art pretrained model now requires substantial computational resources which demands considerable energy, along with the associated financial and environmental costs. 
\citet{vaswani2017attention} report that a Transformer-base model was trained on 8 Nvidia P100 GPUs for 12 hours
and \citet{strubell2019energy} report a BERT-base model was trained on 64 V100 GPUs for 79 hours.  
Thus for alignment attention, we utilize it to finetune these pretrained models on large corpora, which is not only computationally and financially friendly but also accessible to more researchers.

\subsubsection{In-domain language understanding evaluation 
}\label{sec:indomain}
We first evaluate the alignment attention for in-domain language tasks where the training and testing data are from the same domain. 
We conduct experiments on eight benchmark datasets from General Language Understanding Evaluation (GLUE) \cite{wang2018glue} and two Stanford Question Answering Datasets (SQuAD) \cite{rajpurkar2016squad,rajpurkar2018know}. 
Our experiments are based on the state-of-the-art pretrained model, ALBERT \cite{lan2019albert}, a memory-efficient version of BERT \cite{devlin2018bert} with parameter sharing and embedding factorization. Based on Huggingface
PyTorch Transformer \cite{wolf2019transformers}, our implementation uses the base version of ALBERT following the same  setting  from \citet{lan2019albert}. 

\begin{table}[htp!]\vspace{-5mm}
\caption{Performance of alignment attention on GLUE and SQuAD benchmarks.}\vspace{-1mm}
\label{tab:nlp}
\begin{center}
\begin{small}
\begin{sc}
\resizebox{0.97\columnwidth}{!}{
\begin{tabular}{@{}ccccccccccccc@{}}\toprule
 & MRPC & CoLA & RTE & MNLI & QNLI & QQP & SST & STS & SQuAD 1.1 & SQuAD 2.0 \\ \midrule
ALBERT-base & 86.5 & 54.5 & 75.8 & 85.1 & 90.9 & 90.8 & 92.4 & 90.3& 80.86/88.70& 78.80/82.07\\
ALBERT-base+{\bf AA-GAN} & 87.5 & 55.7 & {\bf 77.3} & 85.8 & {\bf 91.3}& 91.4 & 92.6&91.1 & 81.19/88.92 & 79.25/82.57 \\
ALBERT-base+{\bf AA-OT} & 87.9 & 54.6 & 77.0 & 85.7 & 91.2& 91.3 & 92.8&91.2 & 81.13/88.89 & 79.18/82.48 \\
ALBERT-base+{\bf AA-CT} & {\bf 88.6} & {\bf 55.9}  & 77.2 & {\bf 85.9} &  {\bf 91.3} & {\bf 91.5} & {\bf 93.1} & {\bf 91.5} & {\bf 81.32/89.02} & {\bf 79.33/82.71} \\
\bottomrule
 \end{tabular}}
\end{sc}
\end{small}
\end{center}
\vspace{-3mm}
\end{table}
\textbf{Results. } In Table \ref{tab:nlp},
we present the soft attention and the alignment attention (AA) with GAN, optimal transport, and CT, resuming from the same checkpoints. The mean accuracies are reported with 5 independent runs (see full results with the error bars in Table \ref{tab:nlp_full_with_error} in the Appendix). Alignment attention outperforms the soft attention, which indicates that matching the distributions of key and query gives better performance than soft attention and the results are not sensitive to the choice of alignment methods.
Due to the expensive computation for distance measure at each step of OT, we will focus on GAN and CT as alignment methods in the following experiments.
Overall, alignment attention improves the soft attention in both GLUE and SQuAD datasets even by only using alignment attention at the finetuning stage. 

\subsubsection{Generalization across domains}\label{sec:outdomain}
In real applications, it is very common to deploy a neural network model into a new domain with data unseen during training. The model's generalization has been extensively studied in the machine learning community.  Significant past work has studied cross-domain robustness using sentiment analysis \citep{chen2018adversarial, peng2018cross, miller2019simplified}. The recent work from \citet{desai2020calibration} has explicitly elected tasks where out-of-domain performance is substantially lower
and challenging domain shifts are exhibited. Following the setting in \citet{desai2020calibration}, we test the generalization ability of our alignment attention. 
For our in-domain and out-of-domain datasets, we split the development set in half to obtain a held-out, non-blind test set. We conduct experiments on three tasks: (1) \textit{Natural Language Inference.} The Stanford Natural Language Inference (SNLI) corpus is a large-scale entailment dataset \cite{bowman2015large} as the in-domain data. Multi-Genre Natural Language Inference (MNLI) \cite{williams2018broad} can be used as unseen out-of-domain test dataset.
(2) \textit{Paraphrase Detection.} Quora Question Pairs (QQP) is used as in-domain data which includes sentence pairs from Quora that are
semantically equivalent \cite{iyer2017first}. TwitterPPDB (TPPDB) \cite{lan2017continuously} is considered as out-of-domain data.
(3) \textit{Commonsense Reasoning.} Situations With Adversarial Generations (SWAG) is a grounded commonsense reasoning task \cite{zellers2018swag}. The out-of-domain data is HellaSWAG (HSWAG), which is a more challenging benchmark \cite{zellers2018swag}.
We report accuracy and expected calibration error (ECE) for both in-domain(ID) and out-of-domain(OD). ECE is calculated as a weighted average of the difference between each bin’s accuracy and confidence: ECE$:= \sum_i \frac{G_i}{N} |\text{acc}(G_i) - \text{conf}(G_i)|$, where $G_i$, acc$(G_i)$, and conf$(G_i)$ are the count, accuracy, and confidence of samples in the $i$'th group, respectively. The number of groups is $10$ as in \cite{desai2020calibration}.

\begin{table}[htp!]
\vspace{-5.5mm}
\caption{\small Results of domain generalization. We report the accuracy and ECE of various models on both in-domain data and out-of-domain data for three tasks: natural language inference, paraphrase detection, and commonsense reasoning.}\vspace{0mm}
\label{tab:out_of_domain}
\begin{center}
\begin{sc}
\resizebox{0.65\columnwidth}{!}{
\begin{tabular}{@{}lccccc@{}} \Xhline{3\arrayrulewidth}
\small  & \multicolumn{2}{c}{\small Accuracy $\uparrow$} & 
& \multicolumn{2}{c}{\small ECE $\downarrow$}\\
\cmidrule{2-3} \cmidrule{5-6}
\small  & \small ID & \small OD && \small ID  & \small OD  \\ \setlength\arrayrulewidth{2pt}
\small {Natural Language Inference} &SNLI&MNLI&&SNLI&MNLI \\ \midrule
\small DA \citep{parikh2016decomposable} & 84.63\small & 57.12 && {\bf1.02} & 8.79 \\
\small ESIM \citep{chen2017enhanced}& 88.32\small & 60.91 && 1.33 & 12.78 \\
\small BERT-base \citep{desai2020calibration} & 90.04\small & 73.52 && 2.54 & 7.03 \\
\small BERT-base+{\bf AA-GAN} & 90.59 & 74.15 && 2.02 & 5.82 \\
\small BERT-base+{\bf AA-CT} & {\bf90.65} & {\bf74.23} && 1.89 & \bf{5.65} \\ \midrule
\small RoBERTa-base  & 91.23\small & 78.79 && {\bf1.93} & 3.62 \\
\small RoBERTa-base+{\bf AA-GAN} & {91.52} & 79.55 && 2.70 & 3.31\\
\small RoBERTa-base+{\bf AA-CT} & {\bf91.68} & {\bf79.60} && 2.52 & {\bf2.79}\\\Xhline{3\arrayrulewidth}
\small {Paraphrase Detection}& QQP &Twitter & &QQP & Twitter \\ \midrule
\small DA \citep{parikh2016decomposable} & 85.85\small & 83.36 && 3.37 & 9.79 \\
\small ESIM \citep{chen2017enhanced}& 87.75\small & 84.00 && 3.65 & 8.38 \\
\small BERT-base \citep{desai2020calibration}& 90.27\small & 87.63 && 2.71 & 8.51 \\
\small BERT-base+{\bf AA-GAN} & {\bf 90.80} & {\bf 88.34} && {\bf 1.45} & {\bf 7.48} \\
\small BERT-base+{\bf AA-CT} & {90.62} & {88.25} && {1.74} & {7.52} \\ \midrule
\small RoBERTa-base \citep{desai2020calibration}& 91.11\small & 86.72 && 2.33 & 9.55 \\
\small RoBERTa-base+{\bf AA-GAN} & {\bf 91.66} & {87.28} && {\bf 1.78} & {9.45}\\
\small RoBERTa-base+{\bf AA-CT} & {91.53} & {\bf87.33} && {1.89} & {\bf9.40}\\\Xhline{3\arrayrulewidth}
\small {Commonsense Reasoning} & SWAG & HSWAG&&SWAG&HSWAG \\ \midrule
\small DA \citep{parikh2016decomposable} & 46.80\small & 32.48 && 5.98 & 40.37 \\
\small ESIM \citep{chen2017enhanced}& 52.09\small & 32.08 && 7.01 & 19.57 \\
\small BERT-base \citep{desai2020calibration}& 79.40\small & 34.48 && 2.49 & 12.62 \\
\small BERT-base+{\bf AA-GAN} & {79.56} & {35.90} && {1.95} & {12.11} \\
\small BERT-base+{\bf AA-CT} & {\bf 79.60} & {\bf36.25} && {\bf1.86} & \bf{11.78} \\ \midrule
\small RoBERTa-base \citep{desai2020calibration}& 82.45\small & 41.68 && 1.76 & 11.93 \\
\small RoBERTa-base+{\bf AA-GAN} & {83.03} & {42.51} && {1.61} & {9.97}\\
\small RoBERTa-base+{\bf AA-CT} & {\bf83.14} & {\bf42.88} && {\bf1.43}& {\bf9.77}\\
\Xhline{3\arrayrulewidth}
\end{tabular}}
\end{sc} 
\end{center}\vspace{-3mm}
\end{table}

\textbf{Results. } In Table \ref{tab:out_of_domain}, we include open-source implementations of Decomposable Attention (DA)  \citep{parikh2016decomposable} and Enhanced Sequential Inference Model (ESIM) \citep{chen2017enhanced} as baselines. For pretrained models, we
use BERT-base-uncased  \cite{devlin2018bert} and RoBERTa-base \cite{liu2019roberta} from HuggingFace Transformers \cite{wolf2019transformers}. We incorporate the alignment attention in both pretrained models.
Alignment attention consistently outperforms the corresponding soft attention not only for in-domain, confirming our results in Section \ref{sec:indomain}, but also for out-of-domain. The relevantly larger gains on the out-of-domain setting indicate that alignment attention has the better generalization ability across domains. The improved results of ECE demonstrate the better-calibrated model for uncertainty estimation with alignment attention.

\subsubsection{Robustness towards adversarial attacks}\label{sec:adv}
Machine learning models recently have been found vulnerable to adversarial examples that
are legitimate input altered by small and often imperceptible perturbations \citep{goodfellow2014explaining}. 
Therefore, it becomes increasingly important to exam the model's robustness against adversarial attacks. Our alignment
attention imposes a regularization to ensure well-aligned key and query distributions so that it is expected to become more robust to the generated perturbation that would fool the model.
We follow the same settings from Section \ref{sec:indomain} to test the robustness of finetuned ALBERT-base models with soft attention or alignment attention. 
We utilize the TextAttack \cite{morris2020textattack} framework and apply three state-of-the-art untargeted black-box adversarial attacks (1) Textfooler \citep{jin2020bert}: counter-fitted word
embedding swap; (2) Textbugger \citep{li2019textbugger}:  character-level insertion, deletion, swap, and substitution; (3) BAE \citep{garg2020bae}: generating BERT masked token
prediction. 
$1000$ adversarial attacks are conducted for each model with a maximum sentence length of $512$. We report the percentages of failed adversarial attacks in Table \ref{tab:adv}. Higher percentages indicate more robust models.
\begin{table}[htp]
\vspace{-5mm}
\caption{Results of pretrained large-scale models' robustness against adversarial attacks. We report the percentages of failed attacks under three adversarial attacks respectively. }
\label{tab:adv}
\begin{center}
\begin{sc}
\resizebox{0.65\columnwidth}{!}{
\begin{tabular}{@{}llcccccc@{}}\toprule
      {Attack} & {Attention} & {MRPC} & {CoLA} & {RTE} & {QQP} & {SST-2} & {Avg.}  \\ \midrule
\multirow{3}{*}{Textfooler} & base & 6.5 & 2.6 & 16.2 & 25.4 &7.0 & 11.5 \\ 
       & {\bf AA-GAN} & 8.4 & {\bf 7.1} & 14.2 & 31.2 & {\bf 14.5} & 15.1\\
       & {\bf AA-CT} & {\bf8.7} & 6.9 & {\bf 15.3} & {\bf 32.1} & 13.9 & {\bf 15.4}  \\
      \midrule
      \multirow{3}{*}{Textbugger} & base & 10.6 & 16.8 & 19.9 & 30.1 & 40.1 & 23.5 \\ 
       & {\bf AA-GAN} & 12.6 & {\bf 22.4} & 20.7 & 36.0 & 56.1 & 29.6\\ 
       & {\bf AA-CT} & {\bf 13.1} & 20.9 & {\bf 21.0} & {\bf 36.5} & {\bf 57.6} & {\bf 29.8} \\
      \midrule
      \multirow{3}{*}{BAE} & base & 44.8 & 4.9 & 35.6 & 48.8 & 13.9 & 29.6\\ 
       & {\bf AA-GAN} &  44.2 & {\bf 7.3} & 35.8 & 46.5 & {\bf 17.8} & 30.3\\
       & {\bf AA-CT} & {\bf 45.3} & 6.7 & {\bf 36.3} & {\bf 49.4} & 17.5 & {\bf 31.0} \\
\bottomrule
\end{tabular}}
\end{sc}
\end{center}
\vspace{-5mm}
\end{table}

{\bf Results.} The alignment attention shows consistent improvements over the soft attention across all three attacks and achieves significant gains on the average failure rates. The alignment attention demonstrates its robustness and gets along with our intuition that the alignment attention can learn better and more robust key and query distribution due to the use of distributional matching.

\subsection{Alignment attention in graph neural networks}

To test the general applicability of our alignment attention, we also experiment our method with graph attention networks (GAT) \cite{velivckovic2017graph}, where the graph structure is injected into the attention masks in which nodes are able to attend over their neighborhoods’ features in the graph. Leveraging masked self-attentional layers, GAT processes the node features for the node classification.

{\bf Experimental Setup.}
Following the setting in GAT \cite{velivckovic2017graph}, we conduct experiments on three standard
citation network benchmark datasets--- Cora, Citeseer and Pubmed \cite{sen2008collective}--- in a transductive setting, indicating all nodes from training and test are on the same graph \cite{yang2016revisiting}. The details of three datasets and experimental settings are deferred to Appendix~\ref{sec:app_exp}.

\begin{table}[htp]
\vspace{-4mm}
\caption{Classification accuracy for graphs. }
\label{tab:gat_accuracy}
\centering
\resizebox{0.43\columnwidth}{!}{
\begin{tabular}{@{}llllll@{}}\toprule
Attention & Cora & Citeseer & PubMed\\ \midrule
GAT & 83.00 & 72.50 & 77.26 \\
{\bf AA-GAN}  & 83.78\small{$\pm0.2$} & 73.32\small{$\pm0.1$} & 78.77\small{$\pm0.2$}\\
{\bf AA-CT} & {\bf83.80}\small{$\pm0.3$} & {\bf73.49}\small{$\pm0.2$}& {\bf78.79}\small{$\pm0.2$} \\
\bottomrule
\end{tabular}}
\end{table} \vspace{-2mm}
{\bf Results.}
In Table \ref{tab:gat_accuracy}, we report the mean classification accuracies on test nodes over $5$ random runs, 
and the standard deviations of alignment attention.
We experiment with both \ours-GAN and \ours-CT. Table \ref{tab:gat_accuracy} shows that alignment attention consistently improves upon the corresponding baseline models across all three datasets, which further confirms the efficient structure of this alignment attention.  The CT-based alignment attention performs better than GAN-based alignment attention.

\subsection{Attention in visual question answering}\label{sec:vqa}
Visual question answering (VQA) \cite{goyal2017making} is a multi-modal learning task where the model predicts the  
answer given a question conditioning on the image. Self-attention architectures in MCAN \cite{yu2019deep} has been recently proposed to learn the fine-grained semantic meaning of both the image and question. We adapt the proposed alignment attention to MCAN  and compare with soft attention. 
We conduct experiments on the VQA-v2 dataset \cite{goyal2017making} and follow  
the hyperparameters and other settings from \citet{yu2019deep}.  In addition, to investigate the model’s robustness to noise, we construct the noisy dataset by incorporating the Gaussian noise (mean 0, variance 1) to image features \cite{larochelle2007empirical, fan2020bayesian}. Four-layer encoder-decoder based MCAN is used as the baseline model with the soft self-attention.
For each experiment, we report the accuracy on both the original data and the noisy data. As in \citet{fan2020bayesian} and \citet{zhang2021bayesian}, we also report the Patch Accuracy vs Patch Uncertainty (PAvPU) \cite{fan2020bayesian,mukhoti2018evaluating} as a measure of the uncertainty estimation where the $p$-value threshold is set to be $0.05$ and the number of attention weight samples is $20$. Please refer more detailed experimental settings in Appendix~\ref{sec:app_exp}.
 
\begin{table}[htp]
\vspace{-4mm} 
\caption{\small Accuracies and PAvPUs of different attentions on both the original VQA-v2 dataset and the noise ones.}
\label{tab:vqa_accuracy}
\begin{center}
\begin{sc}
\resizebox{0.6\columnwidth}{!}{
\begin{tabular}{@{}lccccc@{}}\toprule
\small  & \multicolumn{2}{c}{\small Accuracy $\uparrow$} & 
& \multicolumn{2}{c}{\small PAvPU $\uparrow$}\\
\cmidrule{2-3} \cmidrule{5-6}
\small  & \small Original & \small Noisy && \small Original  & \small Noisy  \\ \midrule
\small Base & 66.74\small & 63.58 && 71.96 & 68.29 \\
\small {\bf AA-GAN} & 66.92\small & 64.28 && 72.17 & 69.80 \\
\small {\bf AA-CT} & {\bf67.01}\small{$\pm0.02$}& {\bf64.57}\small{$\pm0.03$} && {\bf72.21}\small{$\pm0.03$} & \bf{69.98}\small{$\pm0.04$}\\
\bottomrule
\end{tabular}}
\end{sc} \vspace{-3mm}
\end{center}
\end{table}

\textbf{Results.} 
The results are summarized in Table~\ref{tab:vqa_accuracy}. We report the accuracy and uncertainty of different attentions on both original and noisy data.  In terms of accuracy, alignment attention shows consistent improvements over the soft attention on both original and noisy data. These results verify our conjecture that the alignment attention is more robust to the noise which aligns with our results on adversarial robustness in Section \ref{sec:adv}. For uncertainty, we observe that on both original and noisy data, alignment attention has better uncertainty estimations, meaning that alignment attention in general is more certain on its correct predictions and more uncertain on its mistakes. 

\begin{figure}[htp] 
\vspace{-2mm}
\centering
\includegraphics[width=8cm]{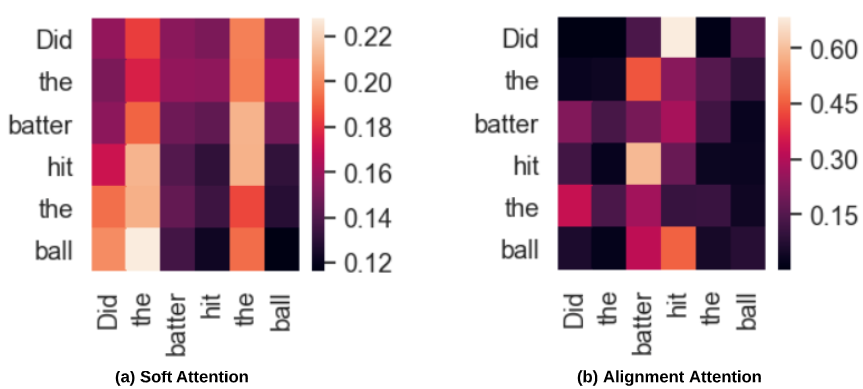}\vspace{-2mm}
\caption{\small For one question from VQA, we visualize the attention weights of {\ours} (a) and soft attention (b). Rows represent queries, and columns represent keys. For example, on the left plot, when the row is `Did' and the column is `hit', the color represents the average attention weight from the query 'Did' to the key `hit'.  
}
\label{fig:att_vis1} \vspace{-4mm}
\end{figure}

\paragraph{Results Analysis. }
\textit{Visualizations.} We plot the attention weights of both alignment attention and soft attention on one question for VQA. In Figure \ref{fig:att_vis1}, attention weight represents the average importance of each query and key pair. For example, in (b), when the row is ‘Did’ and the column is ‘hit’, the color represents the average attention weight from the query `Did’ to the key ‘hit’. We observe that the alignment attention gives relatively sharper attention weights compared to the soft attention
and therefore gives good prediction accuracy and uncertainty estimation.

\paragraph{Parameter Size and Running time.} In Table~\ref{tab:act_parameter}, we provide the parameter sizes and step time for different attention types combined with MCAN
where the attention module constructs the main model. It shows that alignment attention ({\ours}) keeps
the parameter size at the very similar level as soft attention while moderately increasing the step time compared to soft attention. 
\begin{table}[htp]
\vspace{-4mm}
\caption{\small Efficiency on VQA task.}
\label{tab:act_parameter}
\begin{center}
\begin{sc}
\resizebox{0.33\columnwidth}{!}{
\begin{tabular}{@{}lccc@{}}\toprule
Attention & Params $\downarrow$ &s/step $\downarrow$ \\ \midrule
Base  & 43.3M & 0.25\\
{\bf AA-GAN} & 43.4M & 0.31\\
{\bf AA-CT} & 43.4M & 0.36\\
\bottomrule
\end{tabular}}
\end{sc}\vspace{-5mm}
\end{center}
\end{table}

\paragraph{Ablation Study.} We conduct ablation study with AA+CT to exam the role of the alignment-weight hyperparameter $\lambda$ in Equation \ref{eq:overall_loss_fuction} by turning it from $0.01$ to $1$. We find that the experimental results are not sensitive to the choice of the value of the $\lambda$. Any number from $0.01$ to $1$ would give similar results. In all experiments considered in the paper, which cover various noise levels and model sizes, we have simply fixed it as $0.01$. Please see detailed results in Table~\ref{tab:vqa_ablation} in Appendix.

\section{Conclusion}\label{sec:conclusion}
Our proposed alignment attention aims to match the key and query distributions. 
We leverage different alignment methods with a generic and efficient architecture design which requires surprisingly few modifications to standard soft attention and enables us to easily convert existing soft attention models, including pretrained ones, to alignment attention. Our experiments on a variety of language understanding tasks show that alignment attention achieves strong performance in accuracy, uncertainty estimation, domain generalization, and adversarial robustness even by only adding the alignment loss during the finetuning stage. In the real-life scenarios, the attention models have been deployed in many machine learning systems, such as self-driving \cite{kim2017interpretable} and healthcare \cite{choi2016retain}. However, the data from the real practice is biased and long-tailed. Therefore, we see opportunities of our method that can mitigate the risks with uncertainty estimation. Further, on graph node classification and visual question answering, alignment attention demonstrates its general applicability and effectiveness of each component of the proposed structure, showing the great potential to be added as a plug-and-play component 
to many existing attention models.


\section*{Acknowledgements}
The authors acknowledge the support of Grant IIS-1812699
from the U.S. National Science Foundation, the APX 2019 project sponsored by the Office of the Vice
President for Research at The University of Texas at Austin, the support of a gift fund from ByteDance Inc., and the Texas Advanced Computing Center
(TACC) 
for providing HPC resources that have contributed to
the research results reported within this paper.

\small
\bibliographystyle{unsrtnat}
\bibliography{reference.bib}

\clearpage
\appendix
\section{Broader impact} \label{appendix:broaderimpact}
Attention modules have been demonstrated the effectiveness in start-of-the-art neural network models. Our proposed method shows the improvements on five representative tasks indicating its efficacy and general applicability. We hope that our work will encourage the community to pay more attention to key and query distributions in existing attention networks. In real-life scenarios, the attention models have been deployed in many machine learning systems, such as self-driving \cite{kim2017interpretable} and healthcare \cite{choi2016retain}. However, the data from the real practice is often biased and long-tailed. The gap between the training data and testing data might be large. Therefore, an undue trust in deep learning models by incautious usage or imprecise interpretation of model output might lead to unexpected false consequences. Also, with computational consumption, environment sustainable and users friendly are considered. 
Therefore, we see opportunities of our method that can mitigate the risks with uncertainty estimation. The model is more certain on its correct predictions and more uncertain on its mistakes where the human-aid is needed in the real-life applications \cite{ovadia2019can}. The proposed method can also be easily incorporated into the finetune stage which requires much less computation.

\section{Experimental details}\label{sec:app_exp}

\subsection{Natural Language Understanding} 

\subsubsection{Model Specifications for In-domain Evaluation}

With parameter sharing and embedding factorization, ALBERT \citep{lan2019albert} is a memory-efficient version of BERT. We use the ALBERT as the pretrained language model for context embeddings. Our experiment is done on the ALBERT-base model with $12$ attention layers and hidden dimension as $768$. The dimension for factorized embedding is $128$.

\subsubsection{Experimental Settings for In-domain Evaluation}
We conduct experiments on eight benchmark datasets from General Language Understanding Evaluation (GLUE) \cite{wang2018glue} and two version of Stanford Question Answering Datasets (SQuAD) \cite{rajpurkar2016squad,rajpurkar2018know}. The 8 tasks in GLUE are Microsoft Research Paraphrase Corpus (MRPC; \cite{dolan2005automatically}), Corpus of Linguistic Acceptability (CoLA; \cite{warstadt2019neural}), Recognizing Textual Entailment (RTE; \cite{dagan2005pascal}), Multi-Genre NLI (MNLI; \cite{williams2017broad}), Question NLI (QNLI; \cite{rajpurkar2016squad}), Quora Question Pairs (QQP; \cite{iyer2017first}), Stanford Sentiment Treebank (SST; \cite{socher2013recursive}), and Semantic Textual Similarity Benchmark (STS;\cite{cer2017semeval}). For SQuAD, we evaluate on both SQuAD v1.1 and SQuAD v2.0. We leverage the pretrained checkpoint as well as the codebase for finetuing provided by Huggingface PyTorch Transformer \cite{wolf2019transformers}. The detailed experiement setting is summarized in the Table~\ref{tab:albert_setting}. To further confirm that the distribution of the keys and queries are well-aligned after training with alignment loss, we use Maximum Mean Discrepancy (MMD) with standard Gaussian kernels to measure key and query distribution discrepancy and compare MMD with or without alignment loss. Aggregating the MMDs across all heads and layers, on Microsoft Research Paraphrase Corpus (MRPC) task, the total MMD with the alignment loss is 0.0038, while that without the alignment loss is 0.057. We have also tried a single MLP (FC-Relu-FC) structure of the discriminator for adversarial training-based alignment and achieved consistent improvements on GLUE data as: MRPC: 87.4, COLA: 55.7, RTE: 77.2, MNLI: 85.5, QNLI: 91.3, SST: 92.5, STS: 91.1.

\begin{table}[htp!]\vspace{-5mm}
\caption{Experimental settings of each task for in-domain pretrained language model (LR: learning rate, BSZ: batch size, DR: dropout rate, TS: training steps, WS: warmping steps, MSL: maximum sentence length).}\vspace{-1mm}
\label{tab:albert_setting}
\begin{center}
\begin{small}
\begin{sc}
\resizebox{0.8\columnwidth}{!}{
\begin{tabular}{@{}ccccccccc@{}}\toprule
& \text { LR } & \text { BSZ } & \text { ALBERT DR } & \text { Classifier DR } & \text { TS } & \text { WS } & \text { MSL } \\ \midrule
\text { CoLA } & 1.00$e^{-5}$ & 16 & 0 & 0.1 & 5336 & 320 & 512 \\
\text { STS } & 2.00$e^{-5}$ & 16 & 0 & 0.1 & 3598 & 214 & 512 \\
\text { SST\minus2 } & 1.00 $e^{-5}$ & 32 & 0 & 0.1 & 20935 & 1256 & 512 \\
\text { MNLI } & 3.00 $e^{-5}$ & 128 & 0 & 0.1 & 10000 & 1000 & 512 \\
\text { QNLI } & 1.00 $e^{-5}$ & 32 & 0 & 0.1 & 33112 & 1986 & 512 \\
\text { QQP } & 5.00 $e^{-5}$ & 128 & 0.1 & 0.1 & 14000 & 1000 & 512\\
\text { RTE } & 3.00 $e^{-5}$ & 32 & 0.1 & 0.1 & 800 & 200 & 512\\
\text { MRPC } & 2.00 $e^{-5}$ & 32 & 0 & 0.1 & 800 & 200 & 512\\
\text { SQuAD v1.1 } & 5.00 $e^{-5}$ & 48 & 0 & 0.1 & 3649 & 365 & 384 \\
\text { SQuAD } v2.0 & 3.00 $e^{-5}$ & 48 & 0 & 0.1 & 8144 & 814 & 512\\
\bottomrule
\end{tabular}}
\end{sc}
\end{small}
\end{center}
\vspace{-3mm}
\end{table}

\begin{table}[htp!]\vspace{-2mm}
\caption{Results of AA on GLUE and SQuAD benchmarks.}\vspace{-1mm}
\label{tab:nlp_full_with_error}
\begin{center}
\begin{small}
\begin{sc}
\resizebox{0.97\columnwidth}{!}{
\begin{tabular}{@{}ccccccccccccc@{}}\toprule
 & MRPC & CoLA & RTE & MNLI & QNLI & QQP & SST & STS & SQuAD 1.1 & SQuAD 2.0 \\ \midrule
ALBERT-base & 86.5 & 54.5 & 75.8 & 85.1 & 90.9 & 90.8 & 92.4 & 90.3& 80.86/88.70& 78.80/82.07\\
ALBERT-base+AA-GAN & 87.5\footnotesize{$\pm$0.3} & 55.7\footnotesize{$\pm$0.5} & {\bf 77.3}\footnotesize{$\pm$0.6} & 85.8\footnotesize{$\pm$0.3} & {\bf 91.3}\footnotesize{$\pm$0.3}& 91.4\footnotesize{$\pm$0.1} & 92.6\footnotesize{$\pm$0.2} &91.1\footnotesize{$\pm$0.2} & 81.19\footnotesize{$\pm$0.1}/88.92\footnotesize{$\pm$0.1} & 79.25\footnotesize{$\pm$0.1}/82.57\footnotesize{$\pm$0.1} \\
ALBERT-base+AA-OT & 87.9\footnotesize{$\pm$0.2} & 54.6\footnotesize{$\pm$0.5} & 77.0\footnotesize{$\pm$0.4} & 85.7\footnotesize{$\pm$0.2} & 91.2\footnotesize{$\pm$0.1}&  91.3\footnotesize{$\pm$0.2} & 92.8\footnotesize{$\pm$0.3}&91.2 \footnotesize{$\pm$0.3}& 81.13\footnotesize{$\pm$0.1}/88.89\footnotesize{$\pm$0.2} & 79.18\footnotesize{$\pm$0.1}/82.48\footnotesize{$\pm$0.1} \\
ALBERT-base+AA-CT & {\bf 88.6}\footnotesize{$\pm$0.4} & {\bf 55.9}\footnotesize{$\pm$0.3}  & 77.2\footnotesize{$\pm$0.3} & {\bf 85.9}\footnotesize{$\pm$0.2} &  {\bf 91.3}\footnotesize{$\pm$0.1} & {\bf 91.5}\footnotesize{$\pm$0.3} & {\bf 93.1}\footnotesize{$\pm$0.2} & {\bf 91.5}\footnotesize{$\pm$0.2} & {\bf 81.32}\footnotesize{$\pm$0.2}/{\bf89.02}\footnotesize{$\pm$0.1} & {\bf 79.33}\footnotesize{$\pm$0.1}/{\bf82.71}\footnotesize{$\pm$0.1} \\
\bottomrule
 \end{tabular}}
\end{sc}
\end{small}
\end{center}
\vspace{-3mm}
\end{table}

\subsubsection{Model Specifications for Domain Generalizations}
We include the results of Decomposable Attention (DA) \citep{parikh2016decomposable} and Enhanced Sequential Inference Model (ESIM) \citep{chen2017enhanced} as baselines from the open-source implementations AllenNLP \cite{gardner2017deep}. Following the setting in \citet{desai2020calibration}, we also include bert-base-uncased \cite{devlin2018bert} and roberta-base \cite{liu2019roberta} as the pretrained baseline models from HuggingFace Transformers \cite{wolf2019transformers}. For BERT, the finetune epoch is $3$, batch size is $32$, learning rate is 2$e^{-5}$, gradient clip is $1.0$, and no weight decay. For RoBERTA, the finetune epoch is 3, batch size is 32, learning rate is 1$e^{-5}$, gradient clip is $1.0$ and weight decay is $0.1$. The optimizer is AdamW \cite{loshchilov2018decoupled}.

\subsubsection{Experimental Settings for Domain Generalizations}
Following the settings in \citet{desai2020calibration}, we test domain generalization on three challenging tasks: (1) \textit{Natural Language Inference.} The Stanford Natural Language Inference (SNLI) corpus is a large-scale entailment dataset \cite{bowman2015large}. 
Multi-Genre Natural Language Inference (MNLI) \cite{williams2018broad} has the similar entailment data across domains. The MNLI can be seen as out-of-domain test dataset.
(2) \textit{Paraphrase Detection.} Quora Question Pairs (QQP) contains semantically equivalent sentence pairs from Quora
 \cite{iyer2017first}. TwitterPPDB (TPPDB) is considered as out-of-domain data which contains the sentence pairs from the paraphrased tweets \cite{lan2017continuously}.
(3) \textit{Commonsense Reasoning.} Situations With Adversarial Generations (SWAG) is a grounded commonsense reasoning task \cite{zellers2018swag}. HellaSWAG (HSWAG) is out-of-domain data which is a more challenging benchmark \cite{zellers2018swag}. 

\subsubsection{Adversarial Robustness}
For the adversarial attack, we follow the setting from \citet{morris2020textattack} and utilize the same models and training procedures as the in-domain natural language understanding. The maximum sentence length is $512$.

\subsection{Graph Neural Networks}
\subsubsection{Model Specifications}
Following the setting in \citet{velivckovic2017graph}, we use the two-layer GAT model. Models are initialized with Glorot initialization \cite{glorot2010understanding} and trained with the cross-entropy loss using the Adam SGD optimizer \cite{kingma2014adam} with an initial learning rate of $0.01$ for Pubmed, and $0.005$ for all other datasets. 

\subsubsection{Detailed Experimental Settings}
We follow the architecture and hyperparameters settings in \citep{velivckovic2017graph}. The number of attention head is $8$ in the first layer computing $8$ features each followed by an exponential linear unit (ELU) \cite{clevert2015fast} nonlinearity. The second layer is a single-head attention for classification. Dropout \cite{srivastava2014dropout, fan2021contextual} is set as $p = 0.6$ and is applied to both layers' input and normalized attention coefficients. In addition, we apply $L2$ regularization with $\lambda = 0.0005$ during training. Pubmed required
slight changes to the architecture. The second layer has $8$ attention heads and the weight $\lambda$ of $L2$ regularization is $0.001$. Early stopping strategy on both the cross-entropy loss and accuracy on the validation nodes are adopted for Cora, Citeseer and Pubmed \cite{sen2008collective}. The patience is $100$ epochs.

\subsection{Visual Question Answering}
\subsubsection{Model Specifications}
We use the state-of-art VQA models, MCAN \cite{yu2019deep} which consists of MCA layers. Two types of attention in the MCA layer are   self-attention (SA) over question and image features and guided-attention (GA) between question and image features. Mult-head structure is included in each MCA layer with the residual and layer normalization components. By stacking multiple MCA layers, MCAN gradually extract the image and question features through the encoder-decoder structure. Four co-attention layers' MCAN is used in our experiment. 

\subsubsection{Experimental Settings} 
We conduct experiments on the VQA-v2 dataset \cite{goyal2017making}, consisting of human-annotated question-answer   pairs for images from the MS-COCO dataset \cite{lin2014microsoft}. The whole dataset is split into the three parts. For training, there are 40k images and 444k QA pairs. For validation, there are 40k images and 214k QA pairs. For testing, there are 80k images and 448k QA pairs. The evaluation is conducted on the validation set as the true labels for the test set are not publicly available \citep{deng2018latent}. For the noisy dataset, we 
perturb the input by adding Gaussian noise (mean 0, variance 1) to the image features \cite{larochelle2007empirical}. We use the same model hyperparameters and training settings in \citet{yu2019deep} as follows:
the dimensionality of input image features, input question features, and fused multi-modal features are set to be $2048$, $512$, and $1024$, respectively. The latent dimensionality in the multi-head attention is $512$, the number of heads is set to $8$, and the latent dimensionality for each head is $64$. The size of the answer vocabulary is set to $N = 3129$ using the strategy in \citet{teney2018tips}. To train the MCAN model, we use the Adam optimizer \citep{kingma2014adam} with $\beta_1 = 0.9$ and $\beta_2 = 0.98$. The base learning rate is set to $\min(2.5te^{-5}, 1e^{-4})$, where $t$ is the current epoch number starting from $1$. After $10$ epochs, the learning rate is decayed by $1/5$ every $2$ epochs. All the models are trained up to $13$ epochs with the same batch size~of~$64$.

\subsubsection{Ablation Study}
\begin{table}[htp!]
\caption{\small Ablation study of alignment-weight hyperparameter on VQA.}
\centering
\label{tab:vqa_ablation}
\begin{sc}
\resizebox{0.5\columnwidth}{!}{
\begin{tabular}{@{}lccccc@{}}\toprule
\small  & \multicolumn{2}{c}{\small Accuracy $\uparrow$} && \multicolumn{2}{c}{\small PAvPU $\uparrow$}\\
\cmidrule{2-3} \cmidrule{5-6}
\small  & \small Original & \small Noisy &&\small Original 
& \small Noisy \\\midrule
\small $\lambda=1$  & 66.98\small & 64.55 && 72.15 & 69.95 \\
\midrule
\small $\lambda=0.1$ & 67.00\small & 64.54 && 72.18 & 69.93\\
\midrule
\small $\lambda=0.01$ & {\bf67.01} & {\bf64.57} && {\bf72.21} & {\bf69.98}\\
\bottomrule
\end{tabular}}
\end{sc} 
\end{table}



\end{document}


\maketitle


\appendix

\section{Proof for Lemma 1}\label{sec:proof_rao_black}
\begin{proof}
\begin{equation}
\begin{split}
 \mbox{KL}(q_{\phiv}(S)||p_\etav(S)) =& \E_{q_{\phiv}(S)}\left[ \sum_{l=1}^L(\log q_{\phiv}(S_l|S_{1:l-1}) - \log p_\etav(S_l|S_{1:l-1})) \right] \\
 =& \sum_{l=1}^L \E_{q_{\phiv}(S)}\left[ \log q_{\phiv}(S_l|S_{1:l-1}) - \log p_\etav(S_l|S_{1:l-1}) \right] \\
 =& \sum_{l=1}^L \E_{q_{\phiv}(S_{1:l-1})}\E_{q_{\phiv}(S_{l}|S_{1:l-1})}\left[ \log q_{\phiv}(S_l|S_{1:l-1}) - \log p_\etav(S_l|S_{1:l-1}) \right] \\
 =& \sum_{l=1}^L \E_{q_{\phiv}(S_{1:l-1})} \underbrace{\mbox{KL}(q_{\phiv}(S_l|S_{1:l-1})||p_\etav(S_l|S_{1:l-1}))}_{\text{analytic}}. \\
\end{split}\label{eq:rao_black_app}
\end{equation}
\end{proof}

\section{Algorithm }
\begin{algorithm}[h!]
 \caption{Bayesian Attention Modules}
 \label{alg:va}
\begin{algorithmic}\small
 \STATE $\thetav, \etav, \phiv \leftarrow$ Initialze parameters, $t\leftarrow 0$, $\rho\leftarrow$ anneal rate
 \REPEAT
 \STATE $\{\xv_i,\yv_i\}_{i=1}^M\leftarrow$ Random minibatch of M datapoints (drawn from full dataset)
 \STATE $\{\epsilonv_i\}_{i=1}^M\leftarrow$ Random samples
 \STATE $\lambda = \text{sigmoid}(t* \rho)$
 \STATE Compute gradients $\frac 1M \nabla_{\thetav, \etav, \phiv}\sum_i 
 {\mathcal{L}}_{\lambda}(\xv_i, \yv_i, \epsilonv_i)$ according to Eq. (4) 
 \STATE Update $\thetav, \etav, \phiv$ with gradients, $t \leftarrow t +1$
 \UNTIL{convergence}
 \STATE {\bf return:} $\thetav, \etav, \phiv$
\end{algorithmic}
\end{algorithm}

\section{Experiment details}\label{sec:app_exp}

\subsection{Graph neural networks}\label{sec:app_graph}
\subsubsection{Model descriptions}
As in \citet{velivckovic2017graph}, we apply a two-layer GAT model. We summarize the graph attention layer 
here. Denote the input node features as $\hv=\{ \hv_1, ..., \hv_N\}$, where $N$ is the number of nodes. Then, the self-attention weights is defined as:

$$\alpha_{ij}^h=\frac{\exp (\text{LeakyReLU}(\av^h[\mathbf{W}^h \hv_i ||\mathbf{W}^h \hv_j]))}{\sum_{k\in \mathcal{N}_i}\exp (\text{LeakyReLU}(\av^h[\mathbf{W}^h \hv_i ||\mathbf{W}^h \hv_k]))},$$
where $\av^h, \mathbf{W}^h$ are neural network weights for head $h$, and $\mathcal{N}_i$ is the set of neighbor nodes for node $i$. $||$ denotes concatenation.

The output $\hv'=\{ \hv_1', ..., \hv_N'\}$ is computed as:
$$\hv_i' = ||_{h=1}^H \sigma (\sum_{j\in \mathcal{N}_i}\alpha_{ij}^h \mathbf{W}^h \hv_j).$$

\subsubsection{Detailed experimental settings}
We follow the same architectural hyperparameters as in \citet{velivckovic2017graph}. The first layer consists of $H = 8$ attention heads computing $8$ features each, and the second layer has a single head attention following an exponential linear unit (ELU) \citep{clevert2015fast} nonlinearity. 
Then softmax is applied to obtain probabilities. During training, we apply $L2$ regularization with $\lambda = 0.0005$. Furthermore, dropout \citep{srivastava2014dropout} with $p = 0.6$ is applied to both layers’ inputs, as well as to the normalized attention coefficients. Pubmed requires
slight changes for hyperparameter: the second layer has $H = 8$ attention heads, and the $L2$ regularization weight is $\lambda = 0.001$. Models are initialized using Glorot initialization \citep{glorot2010understanding} and trained with cross-entropy loss using the Adam SGD optimizer \citep{kingma2014adam} with
an initial learning rate of $0.01$ for Pubmed, and $0.005$ for all other datasets. In both cases we use
an early stopping strategy on both the cross-entropy loss and accuracy on the validation nodes, with a patience of $100$ epochs. Here, we summarize the hyperparameters for BAM, including anneal rate $\rho$ (as in Algorithm~\ref{alg:va}), $\sigma_1$ and $\sigma_2$ for prior Lognormal and posterior Lognormal respectively, $k$ for Weibull distribution, $\alpha, \beta$ for Gamma distribution, hidden dimension for contextual prior $d_\text{mid}$. On Pubmed, we use anneal rate $\rho=0.2$ for all methods. For BAM-LF, $\sigma_1=1\mathrm{E}6$, $\sigma_2=1\mathrm{E}{\minus2}$. For BAM-LC, $\sigma_1=1\mathrm{E}5$, $\sigma_2=1\mathrm{E}{\minus2}$, and $d_\text{mid}=5$. For BAM-WF, $k=10$, $\beta=1\mathrm{E}{\minus8}$, $\alpha=1\mathrm{E}{\minus4}$. For BAM-WC, $k=10$, $\beta=1\mathrm{E}{\minus4}$, and $d_\text{mid}=5$. On Cora, for BAM-LF, $\sigma_1=1\mathrm{E}{15}$, $\sigma_2=1\mathrm{E}{\minus6}$, and $\rho=0.2$. For BAM-LC, $\sigma_1=1\mathrm{E}{15}$, $\sigma_2=1\mathrm{E}{\minus15}$, $\rho=0.1$, and $d_\text{mid}=1$. For BAM-WF, $k=1$, $\beta=1\mathrm{E}{\minus10}$, $\alpha=1\mathrm{E}{\minus15}$, and $\rho=0.2$. For BAM-WC, $k=1$, $\beta=1\mathrm{E}{\minus10}$, $\rho=0.1$, and $d_\text{mid}=1$. On Citeseer, we use anneal rate $0.1$ for all methods. for BAM-LF, $\sigma_1=1\mathrm{E}{15}$ and $\sigma_2=1\mathrm{E}{\minus6}$. For BAM-LC, $\sigma_1=1\mathrm{E}{15}$, $\sigma_2=1\mathrm{E}{\minus5}$, and $d_\text{mid}=1$. For BAM-WF, $k=100$, $\beta=1\mathrm{E}{\minus15}$, and $\alpha=1\mathrm{E}{\minus7}$. For BAM-WC, $k=100$, $\beta=1\mathrm{E}{\minus15}$, and $d_\text{mid}=1$.
\begin{table}\centering
\caption{Basic statistics on datasets for node classification on graphs.}
\label{tab:gat_data}
\begin{sc}
\resizebox{0.49\columnwidth}{!}{
\begin{tabular}{@{}llllll@{}}\toprule
 & Cora & Citeseer & PubMed \\ \midrule
\#Nodes & 2708 & 3327 & 19717 \\
\#Edges & 5429 & 4732 & 44338 \\
\#Features/Node & 1433 & 3703 & 500 \\
\#Classes & 7 & 6 & 3 \\
\#Training Nodes & 140 & 120 & 60 \\
\#Validation Nodes & 500 & 500 & 500 \\
\#Test Nodes & 1000 & 1000 & 1000 \\
\bottomrule
\end{tabular}}
\end{sc} 
\end{table}

\subsection{Visual question answering}\label{sec:app_vqa}
\subsubsection{Uncertainty evaluation via PAvPU}
\label{sec:uncer_pavpu}
We adopt hypothesis testing to quantify the uncertainty of a model's prediction. 
Consider $M$ posterior samples of predictive probabilities $\{\pv_m\}_{m=1}^M$, where $\pv_m$ is a vector with the same dimension as the 
number of classes. 
To quantify how confident our model is about its prediction, we evaluate whether the difference between the probabilities of the first and second highest classes (in terms of posterior means) 
is statistically significant with two-sample $t$-test. 
With the output $p$-values and a given threshold, we can determine whether a model is certain about its prediction. Then, we evaluate the uncertain using the Patch Accuracy vs Patch Uncertainty metric \citep{mukhoti2018evaluating} which is defined as 
 $\mathrm{PAvPU}={\left(n_{a c}+n_{i u}\right)}/{\left(n_{a c}+n_{a u}+n_{i c}+n_{i u}\right)}$, 
where $n_{ac}, n_{au}, n_{ic}, n_{iu}$ are the numbers of accurate and certain, accurate and uncertain, inaccurate and certain, inaccurate and uncertain samples, respectively. Since for VQA, each sample has multiple annotations, the accuracy for each answer can be a number between $0$ and $1$ and it is defined as $\text{Acc}(ans) = \min \{{(\# \text{human that said } ans)}/{3},1\}.$ Then we generalize the PAvPU for VQA task accordingly:
\bas{
\resizebox{0.45\hsize}{!}{$n_{ac} = \sum_{i} \text{Acc}_i \text{Cer}_i,~n_{iu} = \sum_{i} (1-\text{Acc}_i)(1- \text{Cer}_i)$\,},\notag
\resizebox{0.45\hsize}{!}{$n_{au} = \sum_{i} \text{Acc}_i (1-\text{Cer}_i),~n_{ic} = \sum_{i} (1-\text{Acc}_i)(\text{Cer}_i)$\,},
}
where for the $i$th prediction $\text{Acc}_i$ is the accuracy and $\text{Cer}_i \in\{0,1\}$ is the certainty indicator. 

\subsubsection{Model descriptions}
We use the state-of-the-art VQA model, MCAN \citep{yu2019deep}, to conduct experiments. The basic component of MCAN is Modular Co-Attention (MCA) layer. The MCA layer is a modular composition of two basic attention units: the self-attention (SA) unit and the guided-attention (GA) unit, where the SA unit focuses on intra-modal interactions and GA unit focuses on inter-modal interactions. Both units follow the multi-head structure as in \citet{vaswani2017attention}, including the residual and layer normalization components. The only difference is that in GA, the queries come from a different modality (images) than the keys and values (questions). By stacking MCA layers, MCAN enables deep interactions between the question and image features. We adopt the encoder-decoder structure in MCAN \citep{yu2019deep} with six co-attention layers.



\subsubsection{Detailed experimental settings}
We conduct experiments on the VQA-v2 dataset, which is split into the training (80k images and 444k QA pairs), validation (40k images and 214k QA pairs), and testing (80k images and 448k QA pairs) sets. The evaluation is conducted on the validation set as the true labels for the test set are not publicly available
\citep{deng2018latent}, which we need for uncertainty evaluation. For the noisy dataset, we add Gaussian noise (mean $0$, variance $5$) to image features. We follow the hyperparameters and other settings from \citet{yu2019deep}: the dimensionality of input image features, input question features, and fused multi-modal features are set to be $2048$, $512$, and $1024$, respectively. The latent dimensionality in the multi-head attention is $512$, the number of heads is set to $8$, and the latent dimensionality for each head is $64$. The dropout rate is $0.1$. The size of the answer vocabulary is set to $N = 3129$ using the strategy in \citet{teney2018tips}. To train the MCAN model, we use the Adam optimizer \citep{kingma2014adam} with $\beta_1 = 0.9$ and $\beta_2 = 0.98$. The learning rate is set to $\min(2.5t\mathrm{E}{\minus 5}, 1\mathrm{E}{\minus 4})$, where $t$ is the current epoch number starting from $1$. After $10$ epochs, the learning rate is decayed by $1/5$ every $2$ epochs. All the models are trained up to $13$ epochs with the same batch size~of~$64$. To tune the hyperparameters in BAM, we randomly hold out $20\%$ of the training set for validation. After tuning, we train on the whole training set and evaluate on the validation set. For BAM-LF, $\sigma_1=1\mathrm{E}{9}$, $\sigma_2=1\mathrm{E}{\minus9}$, and $\rho=0.2$. For BAM-LC, $\sigma_1=1\mathrm{E}{9}$, $\sigma_2=1\mathrm{E}{\minus9}$, $\rho=0.2$, and $d_\text{mid}=20$. For BAM-WF, $k=1000$, $\beta=1\mathrm{E}{\minus2}$, $\alpha=1\mathrm{E}{\minus3}$, and $\rho=0.2$. For BAM-WC, $k=1000$, $\beta=1\mathrm{E}{\minus6}$, $\rho=0.1$, and $d_\text{mid}=20$.

\subsubsection{More results}

\begin{table}[hbt!]\centering
\caption{Performance comparison among different attention modules on visual question answering.}
\label{tab:vqa_accuracy_complete}
\begin{sc}
\resizebox{0.49\columnwidth}{!}{
\begin{tabular}{@{}lllcll@{}}\toprule
\multicolumn{6}{c}{Accuracy}\\\midrule
Attention & \multicolumn{2}{c}{Original Data} & 
& \multicolumn{2}{c}{Noisy Data}\\
\cmidrule{2-3} \cmidrule{5-6}
& ALL & Y/N / NUM / OTHER && ALL & Y/N / NUM / OTHER \\ \midrule
Soft & 66.95\small & 84.55 / 48.92 / 58.33 && 61.25 & 80.58 / 40.80 / 51.97 \\
BAM-LF & 66.89 & 84.46 / 49.11 / 58.24 && 61.43 & 80.95 / 41.51 / 51.85 \\
BAM-LC & 66.93 & 84.58 / 49.05 / 58.24 && 61.58 & 80.70 / 41.31 / 52.40 \\
BAM-WF & 66.93 & 84.55 / 48.84 / 58.32 && 61.60 & 81.02 / 41.84 / 52.05 \\
BAM-WC & {\bf67.02} & 84.66 / 48.88 / 58.42 && {\bf62.89}\small & 81.94 / 41.90 / 53.96 \\
\bottomrule
\end{tabular}}
\end{sc} 
\begin{sc}
\resizebox{0.49\columnwidth}{!}{
\begin{tabular}{@{}lllcll@{}}\toprule
\multicolumn{6}{c}{Uncertainty}\\\midrule
Attention & \multicolumn{2}{c}{Original Data} & 
& \multicolumn{2}{c}{Noisy Data}\\
\cmidrule{2-3} \cmidrule{5-6}
& ALL & Y/N / NUM / OTHER && ALL & Y/N / NUM / OTHER \\ \midrule
Soft & 70.04 & 83.02 / 56.81 / 63.66 && 65.34 & 78.84 / 49.80 / 59.18 \\
BAM-LF & 69.92 & 82.79 / 56.87 / 63.58 && 65.48 & 79.13 / 50.19 / 59.16 \\
BAM-LC & 70.14& 83.02 / 57.40 / 63.71 && 65.60 & 78.85 / 49.87 / 59.70 \\
BAM-WF & 70.09 & 83.00 / 56.85 / 63.78 && 65.62 & 79.16 / 50.22 / 59.40 \\
BAM-WC & {\bf71.21}\small & 83.95 / 58.12 / 63.82 && \bf{66.75}\small & 80.21 / 51.38 / 60.58 \\
\bottomrule
\end{tabular}}
\end{sc} 
\end{table}

\subsection{Image captioning}\label{sec:app_ic}
\subsubsection{Model descriptions}
We conduct experiments on an attention-based model for image captioning, Att2in, in \citet{rennie2017self}. This model uses RNN as its decoder, and at each step of decoding, image features are aggregated using attention weights computed by aligning RNN states with the image features. Formally, suppose $I_1, ..., I_N$ are image features, $\hv_{t-1}$ is the hidden state of RNN at step $t-1$. Then, the attention weights at step $t$ are computed by:
$\alphav_t= \text{softmax} (\av_t+b_\alpha)$, and $a_t^i = W\text{tanh}(W_{aI}I_i+ W_{ah}\hv_{t-1}+b_a)$, where $W, W_{aI}, W_{ah}, b_\alpha, b_a$ are all neural network weights. Aggregated image feature $I_t= \sum_{i=1}^N \alpha_t^i I_i$ would then be injected into the computation of the next hidden state of RNN $\hv_t$ (see details in \citet{rennie2017self}).

\subsubsection{Detailed experimental settings}
We use the code from \url{https://github.com/ruotianluo/self-critical.pytorch} and conduct our experiments on the MS COCO dataset \citep{lin2014microsoft} that consists of 123,287 images. Each image has at least five captions. We use the standard data split from \citet{karpathy2015deep}, with 113,287 training, 5000 validation, and 5000 testing images. The vocabulary size $V$ is 9488 and the max caption length $T$ is 16. We replace the ResNet-encoded features in \citet{rennie2017self} with boxing box features extracted from a pre-trained Faster-RCNN \citep{ren2015faster} as visual features. Following the original setting in the code base, we use batch size $10$, Adam optimizer with learning rate $5\mathrm{E}\minus 4$, dropout rate of $0.5$ and train $30$ epochs. During training, we use MLE loss only without scheduled sampling or RL loss, neither of which is compatible with our current framework. For testing, we use greedy search to generate sequences. For BAM, we use contextual prior with $d_\text{mid}=10$ and $\rho=1$. For BAM-WC, $k=10$, $\beta=1\mathrm{E}\minus 6$. For BAM-LC, $\sigma_1=1\mathrm{E}3, \sigma_2=0.1$.



\subsection{Neural Machine Translation}\label{sec:app_nmt}
\subsubsection{Model descriptions}
To make comparision with \citet{deng2018latent}, we adopt the LSTM-based machine translation model in that paper.
The model use a bidirectional LSTM to encode source sentence to source representations $\xv_1, ..., \xv_T$. At the step $j$ of decoding, current LSTM state $\tilde{\xv}$ (a function of previous target words $y_{1:j-1}$) is used as query. The attention weights is computed from an MLP between the query and encoded source token representations. Then the aggregated feature is used to produce the distribution over the next target work $y_j$ (see details in \citet{deng2018latent} or see code in \url{https://github.com/harvardnlp/var-attn}).

\subsubsection{Detailed experimental settings}
\label{sec:nmt}
We use the same dataset, IWSLT \citep{cettolo2014report}, as \citet{deng2018latent}. We preprocess the data in the same way: using Byte Pair Encoding over the combined source/target training set to obtain a vocabulary size of 14,000 tokens \citep{sennrich2015neural}. We train on the sequence length up to 125. We use a two-layer bi-dreictional LSTM with 512 units and 768 units for encoder and decoder respectively. In addition, the batch size is $6$, dropout rate is $0.3$, learning rate is $3\mathrm{E}\minus4$ (Adam optimizer). 
For testing, we use beam search with beam size 10 and length penalty $1$ \citep{wu2016google}. For BAM-WC, $k=5$, $\beta=1\mathrm{E}{\minus6}$, $\rho=1$, and $d_\text{mid}=5$. 

\subsection{Pretrained language model}\label{sec:app_plm}
\subsubsection{Model descriptions}
BERT \citep{devlin2018bert} is a state-of-the-art deep bidirectional transformer\citep{vaswani2017attention} model pretrained on large corpora to extract contextual word representations. ALBERT \citep{lan2019albert} improves upon BERT in terms of latency efficiency and performance by using (a) factorized embedding parameterization, (b) cross-layer parameter sharing, and (c) a sentence-order prediction (SOP) loss. Our experiment is done on the ALBERT-base model, which includes $12$ attention layers, each of hidden dimension $768$. The embedding dimension for factorized embedding is $128$. While BERT-base involves $108M$ parameters, ALBERT-base only has $12M$ parameters.

\subsubsection{Detailed experimental settings}
Our experiment includes both the General Language Understanding Evaluation (GLUE) and Stanford Question Answering (SQuAD) Datasets. We evaluate on $8$ tasks from GLUE including Corpus of Linguistic Acceptability (CoLA; \citep{warstadt2019neural}), Stanford Sentiment Treebank (SST; \citep{socher2013recursive}), Microsoft Research Paraphrase Corpus (MRPC; \citep{dolan2005automatically}), Semantic Textual Similarity Benchmark (STS;\citep{cer2017semeval}), Quora Question Pairs (QQP; \citep{iyer2017first}), Multi-Genre NLI (MNLI; \citep{williams2017broad}), Question NLI (QNLI; \citep{rajpurkar2016squad}), and Recognizing Textual Entailment (RTE; \citep{dagan2005pascal}). We evaluate on both SQuAD v1.1 and SQuAD v2.0.
Our code is built on \citet{wolf2019transformers}, which can be found at \url{https://github.com/huggingface/transformers}. We follow the training settings as in \citet{lan2019albert} and summarize them in Table~\ref{tab:albert_setting}. We also include the hyperparameter setting for BAM-WC. We note, as the model is already pretrained so we do not anneal KL term. We pick $\beta=1\mathrm{E}\minus2$ and $d_\text{dim}=5$ for all experiments, as we found the performance is not sensitive to them. We include the $k$ in Table~\ref{tab:albert_setting}.

\begin{table}[t!] 
\caption{Experiment setting for pretrained language model (LR: learning rate, BSZ: batch size, DR: dropout rate, TS: training steps, WS: warmping steps, MSL: maximum sentence length).}\vspace{-1mm}
\label{tab:albert_setting}
\begin{center}
\begin{small}
\begin{sc}
\resizebox{0.9\columnwidth}{!}{
\begin{tabular}{@{}ccccccccc@{}}\toprule
& \text { LR } & \text { BSZ } & \text { ALBERT DR } & \text { Classifier DR } & \text { TS } & \text { WS } & \text { MSL } & $k$ \\ \midrule
\text { CoLA } & 1.00 $\mathrm{E}$\minus05 & 16 & 0 & 0.1 & 5336 & 320 & 512 & 10\\
\text { STS } & 2.00 $\mathrm{E}$\minus05 & 16 & 0 & 0.1 & 3598 & 214 & 512 & 20\\
\text { SST\minus2 } & 1.00 $\mathrm{E}$\minus05 & 32 & 0 & 0.1 & 20935 & 1256 & 512 &1000\\
\text { MNLI } & 3.00 $\mathrm{E}$\minus05 & 128 & 0 & 0.1 & 10000 & 1000 & 512 & 5\\
\text { QNLI } & 1.00 $\mathrm{E}$\minus05 & 32 & 0 & 0.1 & 33112 & 1986 & 512 & 500\\
\text { QQP } & 5.00 $\mathrm{E}$\minus05 & 128 & 0.1 & 0.1 & 14000 & 1000 & 512& 1000\\
\text { RTE } & 3.00 $\mathrm{E}$\minus05 & 32 & 0.1 & 0.1 & 800 & 200 & 512& 1000 \\
\text { MRPC } & 2.00 $\mathrm{E}$\minus05 & 32 & 0 & 0.1 & 800 & 200 & 512& 100\\
\text { SQuAD v1.1 } & 5.00 $\mathrm{E}$\minus05 & 48 & 0 & 0.1 & 3649 & 365 & 384& 10 \\
\text { SQuAD } v 2.0 & 3.00 $\mathrm{E}$\minus05 & 48 & 0 & 0.1 & 8144 & 814 & 512& 2000\\
\bottomrule
\end{tabular}}
\end{sc}
\end{small}
\end{center}
\vspace{-3mm}
\end{table}

\small
\bibliographystyle{plainnat}
\bibliography{reference.bib}
\normalsize

\newpage